\documentclass{article}

\PassOptionsToPackage{numbers, compress}{natbib}

\usepackage[final]{neurips_2021}  

\usepackage[utf8]{inputenc} 
\usepackage[T1]{fontenc}    
\usepackage{hyperref}       
\usepackage{url}            
\usepackage{booktabs}       
\usepackage{amsfonts}       
\usepackage{nicefrac}       
\usepackage{microtype}      
\usepackage{xcolor}         
\usepackage{amsmath}
\usepackage{adjustbox}
\usepackage{mathtools}
\usepackage{multirow}
\usepackage{wrapfig}
\usepackage{graphicx}
\usepackage{caption}
\usepackage{subcaption}
\usepackage{cleveref}

\usepackage{kotex}  

\title{Conditional Generation of Periodic Signals with Fourier-Based Decoder}

\author{%
  Jiyoung Lee \\
  KAIST AI\\
  Daejeon, South Korea \\
  \texttt{jiyounglee0523@kaist.ac.kr} \\
   \And
   Wonjae Kim \\
   NAVER AI LAB \\
   Seongnam-si, South Korea \\
   \texttt{wonjae.kim@navercorp.com} \\
   \AND
   Daehoon Gwak \\
  KAIST AI\\
  Daejeon, South Korea \\
   \texttt{daehoon.gwak@kaist.ac.kr} \\
   \And
   Edward Choi \\
  KAIST AI\\
  Daejeon, South Korea \\
   \texttt{edwardchoi@kaist.ac.kr} \\

}

\usepackage{xcolor}         

\begin{document}

\maketitle

\begin{abstract}
Periodic signals play an important role in daily lives. Although conventional sequential models have shown remarkable success in various fields, they still come short in modeling periodicity; they either collapse, diverge or ignore details. In this paper, we introduce a novel framework inspired by Fourier series to generate periodic signals. We first decompose the given signals into multiple sines and cosines and then conditionally generate periodic signals with the output components. We have shown our model efficacy on three tasks: reconstruction, imputation and conditional generation. Our model outperforms baselines in all tasks and shows more stable and refined results.
\end{abstract}


\section{Introduction}
\vspace{-2mm}

Periodic signals exist in daily lives.
In biomedical domain, electrocardiogram (ECG) \citep{kutlu2016arrhythmia} and body temperature \citep{ishida1999biological, refinetti1992circadian} are critical periodic signals in examining patients health status.
ECG, in particular, is an important measure to diagnose patient heart diseases \citep{heden1996detection} such as Myocardial Infarction, AV Block, and Ventricular Tachycardia.
Despite its significance, most publicly open-sourced ECG datasets \citep{moody2001impact, bousseljot1995nutzung} are small in size and they all tend to have a severe data imbalance problem \citep{shaker2020generalization}--normal or common disease ECG records make up the majority while rare diseases barely exist.
Therefore, generating ECG records conditioned on diagnosis can be beneficial in solving data imbalance issue and further developing ECG deep learning models. 


Sequential models, such as recurrent neural networks (RNNs) \citep{LSTM, GRU}, transformer decoder \citep{transformer}, neural ordinary differential equations (NODEs) \citep{chen2018neuralode}, and neural processes (NPs) \citep{CNP, NP, ANP} have shown excellent results in various fields \citep{seq2seq, GPT3, Dalle}. 
However, as shown in \cref{Experiment}, there is room for improvement when it comes to generating periodic signals.
These models either collapse, diverge or ignore subtle periodic details.
In theory, discrete Fourier series (DFS) or discrete Fourier transform (DFT) can handle sampled periodic signals.
In practice, however, DFS and DFT perform poorly when signals are obtained in irregular timesteps or contain noise \citep{Silvescu, zhang2020periodic}, and are difficult to serve as (conditional) generative models.

In this paper, we introduce a novel architecture that directly utilizes Fourier series for generating periodic signals.
By the definition of Fourier series, periodic signals are composed of sinusoids.
Built upon conditional variational autoencoder \citep{CVAE}, we first obtain latent representation vector of the signal with the encoder.
The decoder then outputs Fourier coefficients, which represent how much each sinusoid contributes to compose the original signal, using the sampled latent vector.
We further generate the label conditioned periodic signals by adding the class information to the sampled latent vector. 
We evaluate our model on three tasks: reconstruction, imputation, and conditional generation to demonstrate its usefulness.
Our model outperforms baseline models in all tasks, suggesting a new direction for modeling periodic data.

\section{Related Work}
\vspace{-2mm}
\textbf{Modeling Periodic Signals \enskip}
Previous research that tried to model periodic signals replaces non-linear activation functions by sinusoids such as $sin(x)$, $cos(x)$ or the linear combination of the two \citep{Silvescu, zhumekenov2019fourier, parascandolo2016taming, snake}.
However, due to the activations' \textit{non-monotonicity}, the periodic activation functions induce numerous local minima, causing troubles in model optimization \citep{parascandolo2016taming}. 

\textbf{Fourier Neural Networks (FNNs) \enskip}
FNNs are neural networks that resemble Fourier series.
\citet{gallant1988there} suggest \textit{cosine squasher} as an activation function. 
\citet{Silvescu} use $cos(x)$ as an activation function to mimic Fourier series.
\citet{liu2013fourier} proposes a combination of $cos(x)$ and $sin(x)$ as activation function, and showed comparable results on various datasets empirically. 
FNN had a superior performance in practical tasks such as aircraft engine fault diagnostics \cite{tan2006fourier}, and control of a class of uncertain nonlinear systems \cite{zuo2005tracking, zuo2008adaptive, zuo2008fourier}.


\textbf{Conditional Time-Series Generation \enskip}
Previous conditional time-series generative work employ convolutional neural networks (CNNs) \citep{borovykh2017conditional, cheng2020attention} and RNNs \citep{esteban2017real}. 
They often make use of generative adversarial networks (GANs) to achieve a realistic-looking data \citep{ramponi2018t}.
\citet{esteban2017real} achieve conditional generation by setting RNNs in both generator and discriminator and concatenating a class label for every time steps. 

\section{Proposed Methods}
\vspace{-2mm}
\begin{figure}[]
  \includegraphics[width=\linewidth]{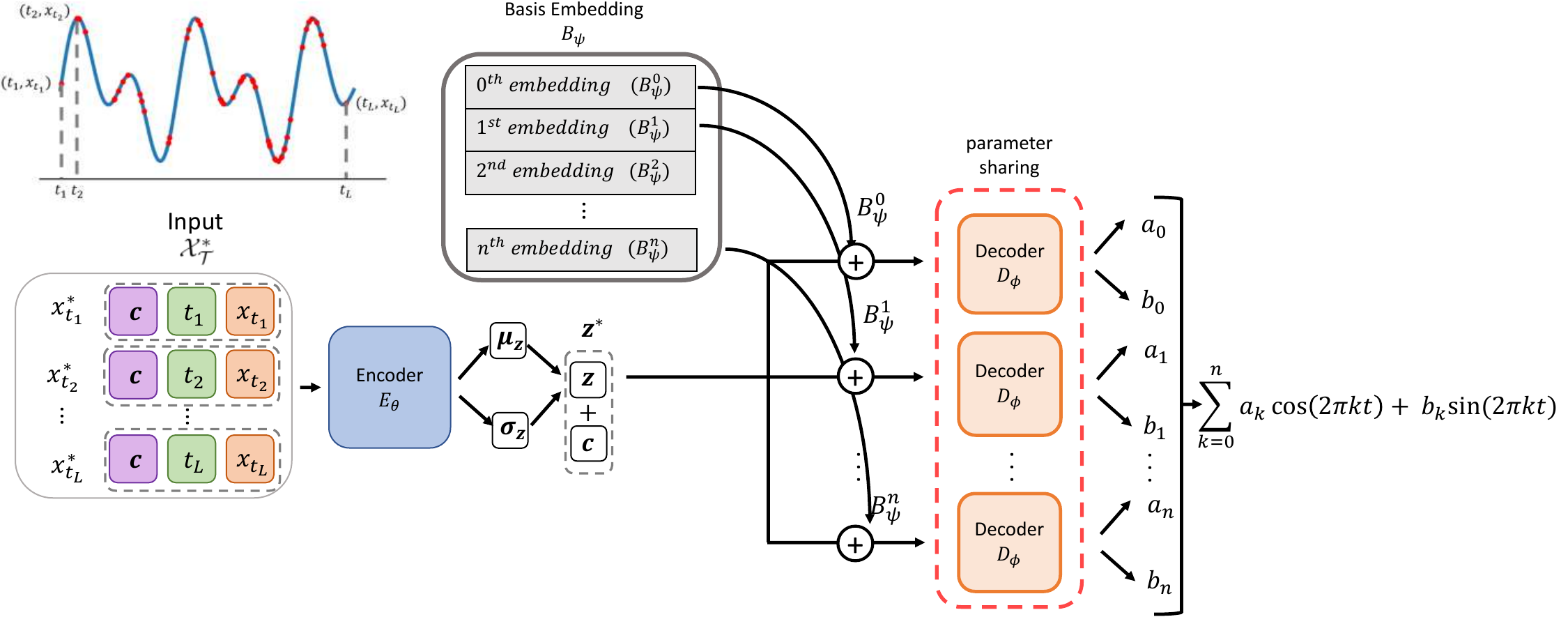}
  \caption{\label{Model}Model architecture. Input $x^{*}_{t_i}$ is a concatenation of input signal $x_{t_i}$, time $t_i$, and label $\mathbf{c}$. We sample $\mathbf{z}$ from $\mathcal{N(\mathbf{\mu}_{\mathbf{z}}, \mathbf{\sigma}_{\mathbf{z}})}$, and concatenate with label $\mathbf{c}$, producing $\mathbf{z^*}$.
  Then $\mathbf{z^*}$ is added with the basis embeddings $B_{\psi}^{k}$ and passed through the decoder $D_\phi$, which produces $a_k$ and $b_k$, namely the \textit{k-th} approximate Fourier coefficients.
  With a set of Fourier coefficients $\mathcal{A} = \{a_0, \ldots, a_n\}$ and $\mathcal{B} = \{b_0, \ldots, b_n\}$, we build an approximated Fourier series.}
  \vspace{-2mm}
\end{figure}

\subsection{Problem Definition}
\vspace{-2mm}
In this section, we will clarify notations to be used throughout the paper and define our task. 
Let $\mathcal{T} = \{ t_1, t_2, \ldots, t_L \}$ be a set of $L$ ordered timesteps with $t_i \in \mathbb{R}$, and $\mathcal{X_T} = { \{x_{t_1}, x_{t_2}, \ldots, x_{t_L}\} }$ be a corresponding input sequence, where $x_{t_i} \in \mathbb{R}$. We note that the intervals among $\mathcal{T}$ can be irregular.
Additionally, the input sequence $\mathcal{X_T}$ has its own one-hot vector label $\mathbf{c} \in \mathbb{R}^{n_c}$, where $n_c$ is the number of unique labels.
Given a label $\mathbf{c}$, the goal of conditional generation is to generate $\mathcal{X_S}$ with another ordered timesteps $\mathcal{S} = \{ s_1, s_2, \ldots, s_M \}$ that maximizes $P(\mathcal{X_S} \ |\ \mathbf{c})$.




\subsection{Background: Fourier Series}
\vspace{-2mm}
Fourier series approximates any periodic function $f(t)$ as an infinite sum of sines and cosines with increasing frequency as written in \cref{fourierseries}. 
Here, $P$ is the period of function $f(t)$. 
\begin{equation}
    f(t) = \sum^{\infty}_{k=0}A_{k}cos\bigg(\frac{2\pi kt}{P}\bigg) + B_{k}sin\bigg(\frac{2\pi kt}{P}\bigg)
    \label{fourierseries}
\end{equation}
$A_k \in \mathbb{R}, B_k \in \mathbb{R}$ are \textit{Fourier coefficients}.
They represent how much each sine and cosine is contributing to formulate the given function.
Hereafter, we will name the set of sines and cosines sharing the same frequency as \textit{basis} and we will assume all signals have a period of 1.
Originally, Fourier coefficients are acquired by integrating the original function $f(t)$ with corresponding sines or cosines.
However, this calculation is infeasible when the data points are discrete and irregular.

For the most of the real signals, finite numbers of basis are sufficient to model a function $f(t)$.
As such, we will predefine the number of basis, denoted as $n$.
In our model, given an input sequence $\mathcal{X_T}$, we approximate the true value of $A_{i}$ and $B_{i}$.
We denote the approximations as $\mathcal{A} = \{a_{0}, \ldots, a_{n}\}$ and $\mathcal{B} = \{b_{1}, \ldots b_{n}\}$: $a_{i}$ and $b_{i}$ approximate $A_{i}$ and $B_{i}$, respectively.

\subsection{Model Architecture}
\vspace{-2mm}
Our model is built upon conditional variational autoencoder \citep{CVAE}.
The overall architecture is illustrated in \cref{Model}.
We construct the input sequence $\mathcal{X^{*}_{T}} = [ {\mathbf{x}^{*}_{t_1}, \mathbf{x}^{*}_{t_2}, \ldots, \mathbf{x}^{*}_{t_L}} ]$ where $\mathbf{x}^{*}_{t_i}$ is a concatenation of $x_{t_i} \in \mathcal{X_T}$, $t_i \in \mathcal{T}$, and label $\mathbf{c}$ as follows:
\begin{equation}
\mathbf{x}^{*}_{t_i} = [x_{t_i} ; t_i ; \mathbf{c}]   
\end{equation}
\textbf{Encoder \enskip}
$E_{\theta}$ produces $\boldsymbol{\mu}_{z} \in \mathbb{R}^z$ and $\boldsymbol{\sigma}_z \in \mathbb{R}^z$ from the input sequence $\mathcal{X^{*}_{T}}$ and we sample the latent variable $\mathbf{z} \in \mathbb{R}^z$ from $q(\mathbf{z}|\mathcal{X^{*}_{T}}) := \mathcal{N}(\boldsymbol{\mu}_z, diag(\boldsymbol{\sigma}_z))$.
$E_\theta$ can be any type of model (\textit{e.g.,} 1-D CNNs, RNNs, transformer encoder) capable of producing an informative representation of the input sequence. 

\textbf{Decoder \enskip}
$\mathbf{z} \in \mathbb{R}^z$ is sampled from $q(\mathbf{z}|\mathcal{X^{*}_{T}})$ and concatenated with $\mathbf{c} \in \mathbb{R}^{n_c}$, producing $\mathbf{z^*} \in \mathbb{R}^{z + n_c}$ as $\mathbf{z^*} = [\mathbf{z} ; \mathbf{c}]$.
We have a basis embedding lookup table denoted as $B_\psi \in \mathbb{R}^{n \times (z+n_c)}$, where $n$ is the predefined number of bases.
The \textit{k-th} row of $B_\psi$, $B^k_\psi$, represents the embedding vector for the \textit{k-th} frequency basis.
In order to obtain the \textit{k-th} Fourier coefficient, $a_k$ and $b_k$, we add $B^k_\psi$ to the latent variable $\mathbf{z^*}$.
Note that this addition is done in parallel, so we can compute it efficiently even when $n$ is large.
The decoder $D_\phi$ produces $a_k$ and $b_k$ as follows:
\begin{equation}
    a_k, b_k = D_\phi(\mathbf{z}^* + B^k_\psi)
\end{equation}
With $\mathcal{A} = \{a_{0}, \ldots, a_{n}\}$ and $\mathcal{B} = \{b_{1}, \ldots b_{n}\}$, we build an approximated Fourier series in \cref{completefourier}.
\begin{equation}
  \hat{f}(t) = \sum^{n}_{k=0}\big(a_{k}cos(2\pi kt) + b_{k}sin(2\pi kt)\big) \label{completefourier}
\end{equation}
Finally, we construct a sequence of predictions as $\widehat{\mathcal{X}}_\mathcal{T} = [\hat{x}_{t_1}, \hat{x}_{t_2}, \ldots, \hat{x}_{t_L}]$ where $\hat{x}_{t_i} = \hat{f}(t_i)$.
We optimize our model by the summation of reconstruction loss between the true observations $\mathcal{X}_\mathcal{T}$ and the predicted observations $\widehat{\mathcal{X}}_\mathcal{T}$, and the Kullback–Leibler divergence loss ($D_{KL}$) between the posterior distribution $q(\mathbf{z}|\mathcal{X^{*}_{T}})$ and the prior distribution  $p(\mathbf{z})=\mathcal{N}(\mathbf{0}, \mathbf{I})$.
The overall loss is:
\begin{equation}
  \mathcal{L} = \sum^{L}_{i=1}\|x_{t_i} - \hat{x}_{t_i}\|_2^2 + \beta D_{KL}(q(\mathbf{z}|\mathcal{X^{*}_{T}}) || p(\mathbf{z})),
\end{equation}
where $\beta > 0$ is a hyperparameter.
After training, to generate $\mathcal{X_S}$ conditioned on label $\mathbf{c}$, we sample $\mathbf{z}$ from the prior distribution $p(\mathbf{z})$ and concatenate with $\mathbf{c}$.
Then we can generate $\mathcal{X_S}$ by decoding the concatenated vector.
Currently, our model assumes all input signals have the period of 1. Extension of this model to dynamically deal with signals of varying periods is our primary future research interest.



\section{Experiments}
\vspace{-2mm}
\label{Experiment}
\subsection{Experimental Setup}
\vspace{-2mm}
We conduct experiments with two periodic datasets: toy sinusoid dataset and Physionet2021 \citep{reyna2021will}. 
Toy dataset is a simple mixture of three sine and cosine functions: $\sum^{3}_{i=1}m_{2i-1}cos(2\pi d_{2i-1}t) + m_{2i}sin(2\pi d_{2i})$.
We put four class conditions for the toy dataset based on amplitudes $\mathcal{M} = \{m_1, \ldots, m_6\}$ and frequencies $\mathcal{D} = \{d_1, \ldots, d_6\}$, resulting in four amplitude-frequency class labels: `Low Amp. \& Low Freq.', `Low Amp. \& High Freq.', `High Amp. \& Low Freq.', and `High Amp. \& High Freq.'. 
`Low Amp.' classes sample $m$ from a uniform distribution $U(1, 4)$ whereas `High Amp.' classes sample $m$ from $U(6, 9)$.
`Low Freq.' classes sample $d$ from $U(1, 4)$, whereas `High Freq.' classes sample $d$ from $U(8, 11)$.
Each signal has a total of 500 timesteps.

Physionet2021 \citep{reyna2021will} contains 12-lead ECG recordings collected from six separate datasets. 
We cropped each record into one second consisting of 500 timesteps and extracted samples with three diagnoses, namely `Right Bundle Branch Block' (RBBB), `Left Bundle Branch Block' (LBBB), and `Atrial Fibrillation' (AF) from the V6 lead.
These diagnoses are selected because they can be examined within one second record \citep{da2002bradycardias}. 
After preprocessing, there were total 36,110 cropped ECG samples.
We split the data into train, validation and test sets with the ratio of 8:1:1. 
More details on preprocessing are in \cref{Appen:DataPreprocess}.

We sampled 20\% of the timesteps during training to insure the irregularity of time.
\footnote{We also conduct experiments without sampling to compare model performance in two scenarios (sparse irregular timeseries VS dense regular timeseries).
The results are reported in \cref{Appen:Experimentwosampling}, where our model outperforms the baselines.
However, our model is particularly more powerful when input sequences are irregular, indicating its usefulness in handling real signals where irregularities exist due to missing timesteps \citep{yang2020missing}.}
For all experiments, we employ 5-layer 1D CNN as an encoder $E_\theta$ and 6-layer MLP as a decoder $D_\phi$.
We evaluate our model on three tasks: (1) reconstruction (for the sampled 20\%), (2) imputation for missing timesteps (for the non-sampled 80\%) and (3) conditional generation.
For the tasks (1) and (2), we use sampled time points and make the model perform both reconstruction and imputation in parallel by generating $\widehat{\mathcal{X}}_\mathcal{T}$ for all timesteps.
For the task (3), we sample $\mathbf{z}$ from the prior distribution $\mathcal{N}(\textbf{0}, \textbf{I})$ and pass through the decoder $D_\phi$. 
With the same 5-layer 1D CNN encoder, we compare our model with four different baseline decoders: Gated Recurrent Units (GRU), Transformer decoder, Neural ODE (NODE) and Neural Processes (NP). Further model implementation details are explained in \cref{Appen:Experiments}.
\begin{wraptable}{R}{0.5\textwidth}
\centering
\vspace{-0.2cm}
\caption{\label{sinimputation} Reconstruction and imputation MSE on Toy Dataset}
\resizebox{5.5cm}{!}{%
\begin{tabular}{ccc}
\toprule
               & Reconstruction & Imputation \\ \hline
GRU            & 72.454         & 74.711     \\
Transformer    & 183.203        & 172.338    \\
NODE          & 15.905         & 19.896     \\
NP             & 3.813          & 4.615      \\ \hline
Fourier (Ours) & \textbf{0.649} & \textbf{0.686} \\
\bottomrule
\end{tabular}%
}
\vspace{-0.2cm}
\end{wraptable}
\begin{figure}[t]
\includegraphics[width=0.9\linewidth]{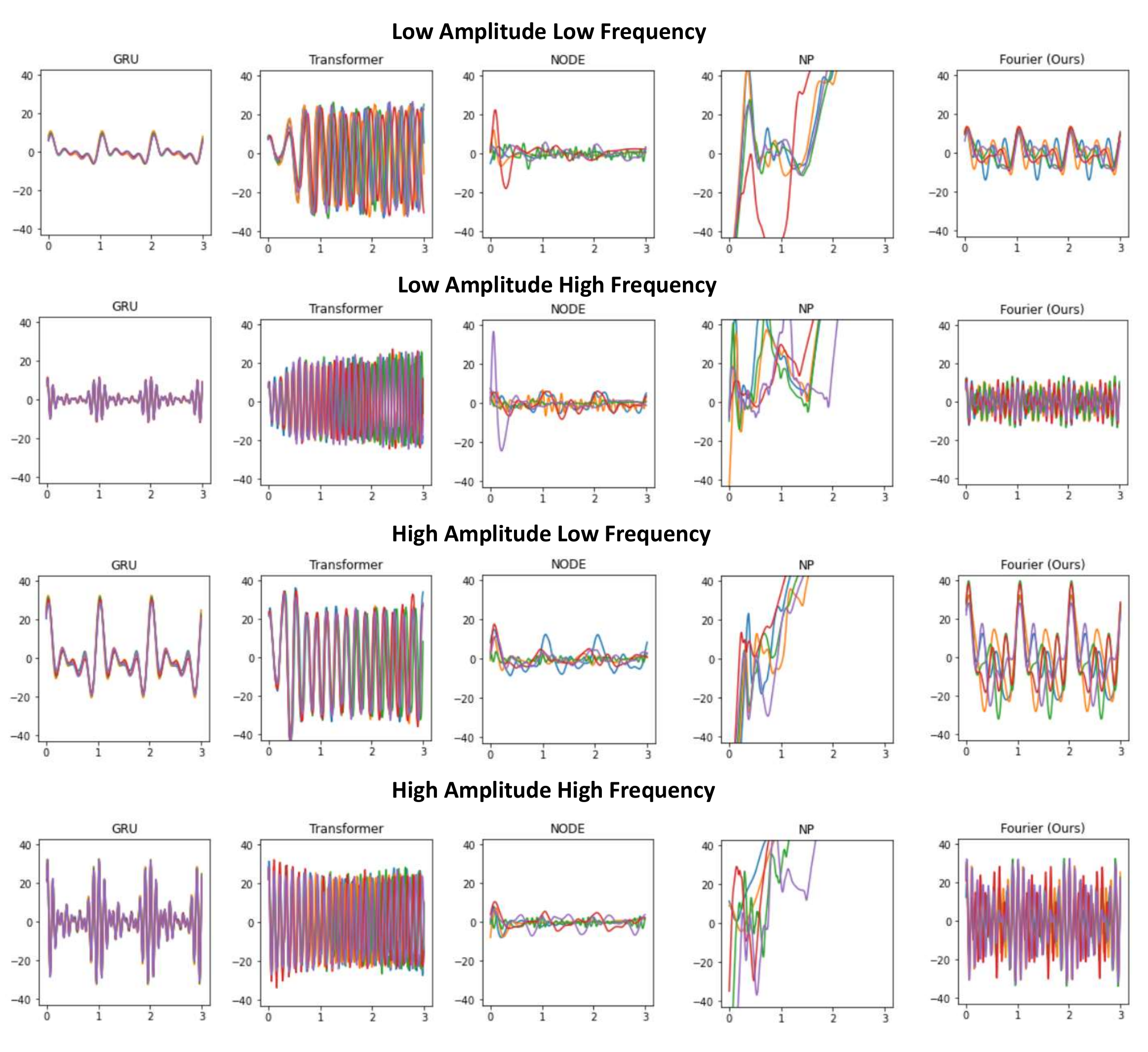}
\caption{Conditionally generated samples in toy dataset. We draw five generated samples for each model plotted in different colors. Our model is able to generate diverse samples while assimilating class conditions whereas baseline models either collapse, diverge, or have a low sample diversity.}
\label{sin_gen_diverse}
\end{figure}
\begin{figure}[t]
\includegraphics[width=0.9\linewidth]{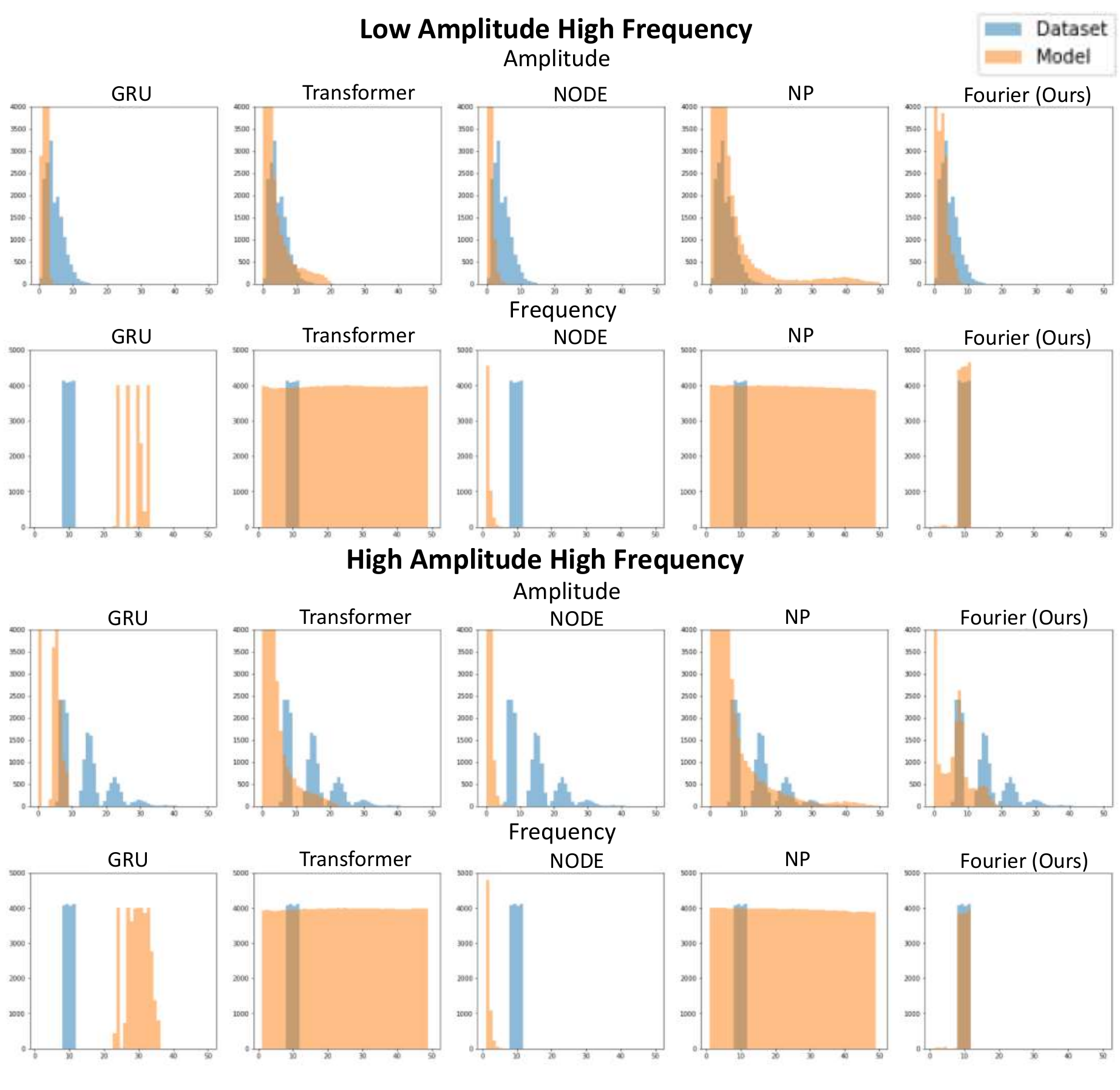}
\caption{Fourier series analysis on conditionally generated samples. We plot histograms on amplitude and frequency for each model from Fourier analysis. Blue color represents the original dataset histogram and orange color represents each model. Note that the Transformer and NP cover more space than original dataset meaning that those models use more sinusoid to generate a single sample than the original dataset.}
\label{sin_hist}
\end{figure}

\subsection{Experiment Results of Toy Dataset}
\vspace{-2mm}
We report reconstruction and imputation results in \cref{sinimputation}.
Our model shows the lowest MSE on both reconstruction and imputation compared to other baseline models.
We illustrate the reconstruction results in \cref{sin_recon} in the \cref{Appen:AdditionalExperiments}.
Based on \cref{sin_recon}, we found both GRU and Transformer to perform very poorly, though GRU was able to capture partial periodicity.
NODE and NP showed better performance on low frequency samples, but they were not able to reproduce subtle details for high frequency samples. 
In contrast, our model showed superior performance across all amplitude-frequency classes.

We conditionally generate 2,000 samples from the sampled $\mathbf{z}$ for each amplitude-frequency class, and for each decoder.
As visualized in \cref{sin_gen_diverse}, NODE produced flattened samples and NP generated non-periodic signals with a large amplitude regardless of the class conditions. 
GRU and Transformer were able to produce periodic signals, but all 2,000 samples were nearly identical with minimal sample diversity. 
In contrast, our model was able to generate diverse periodic signals.  


In order to verify whether the generated samples correctly reflect the class conditions, we conduct Fourier series analysis, which decomposes a given signal into multiple sines and cosines as expressed in \cref{fourierseries}. 
From the analysis, we can calculate which frequency is used to compose the signal and its corresponding coefficients (\textit{i.e.} amplitude).
We plot histograms on both amplitude and frequency for the toy dataset, and compare the baselines with our model.
The results are shown in \cref{sin_hist} for `Low Amplitude \& High Frequency' and `High Amplitude \& High Frequency', the rest two class conditions are described in \cref{sin_his_appen} in the \cref{Appen:AdditionalExperiments}. 
In all baseline models, their amplitude and frequency histograms are similarly shaped across the two classes, which implies that those models fail to reflect the class conditions when generating samples. 
Our model overlaps with the dataset distribution most precisely compared to the other baselines, suggesting that the periodic signals generated by our model are properly conditioned on the class.


\subsection{Experiment Results of Electrocardiogram}
\vspace{-2mm}
\begin{wraptable}{R}{0.5\textwidth}
\centering
\vspace{-0.2cm}




\caption{\label{ECGimputation} Reconstruction and imputation MSE on Physionet2021}
\resizebox{5.5cm}{!}{%

\begin{tabular}{ccc}
\toprule

               & Reconstruction & Imputation \\ \hline
GRU            & 77.393         & 75.327     \\
Transformer    & 68.823         & 35.692     \\
NODEs          & 4.937          & 3.262      \\
NP             & 2.630          & 1.840      \\ \hline
Fourier (Ours) & \textbf{2.164}          & \textbf{1.519} \\
\bottomrule
\end{tabular}%
}
\vspace{-0.2cm}
\end{wraptable}
\begin{figure}
    \centering
    \includegraphics[width=\linewidth]{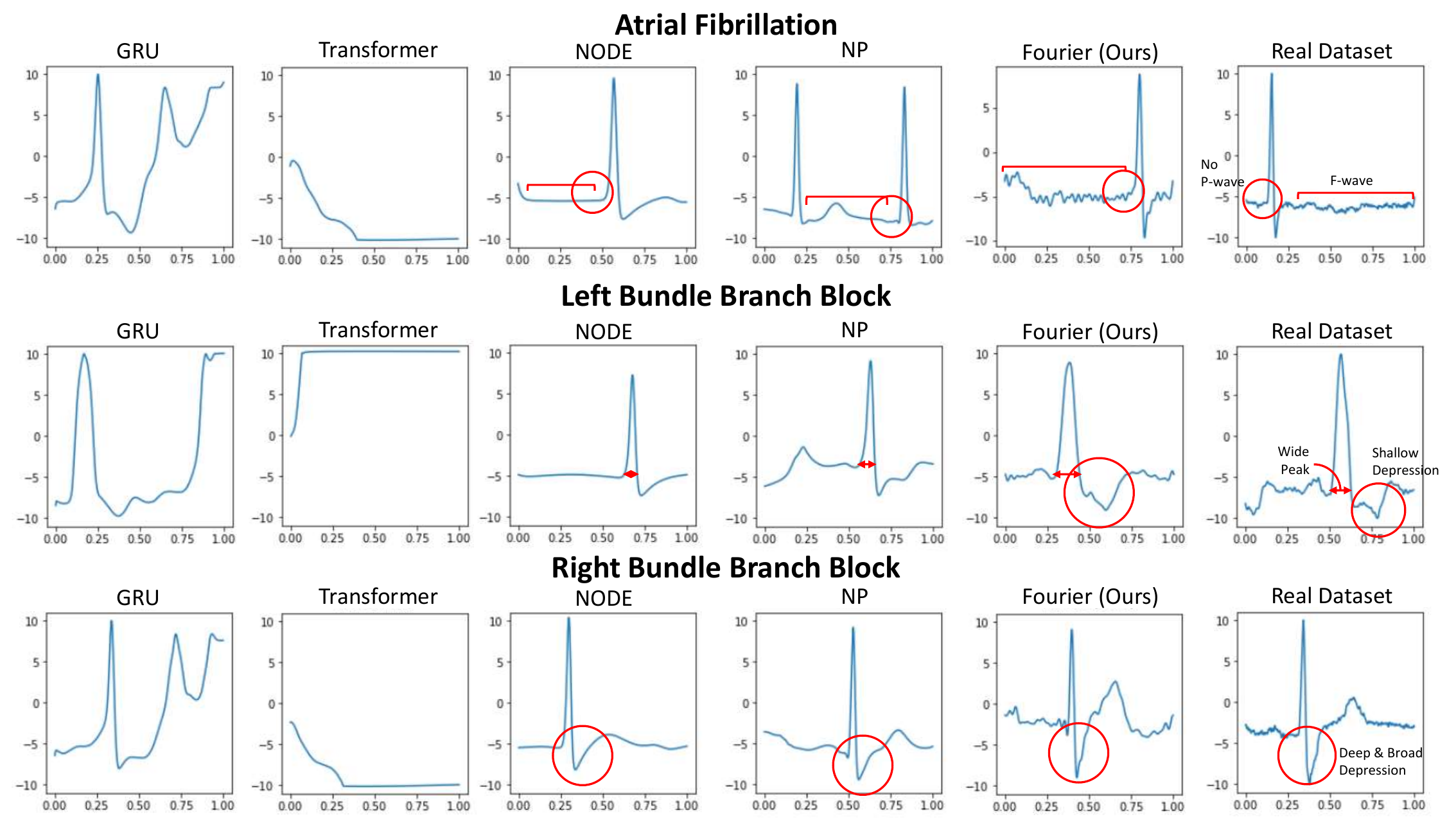}
    \caption{Conditionally generated ECG samples for each diagnosis. Real dataset samples are from Physionet2021. Main characteristics for each diagnosis and corresponding spots in generated samples are marked in red.}
    \label{ECGgen}
\end{figure}
We report reconstruction and imputation results in \cref{ECGimputation}. 
Our model shows the lowest MSE in both reconstruction and imputation. 
The results are visualized by \cref{ECG_recon} in \cref{Appen:AdditionalExperiments}. 
Based on \cref{ECG_recon}, GRU and Transformer cannot capture the peak of given ECG records. 
Although NODE and NP could grasp the peak point, they disregard details such as subtle waves in the isoelectric line, the straight line on the ECG. 
Our model can both capture the peak and the subtle fluctuations clearly.

We conditionally generate 3,000 samples for each diagnosis and decoder.
As visualized in \cref{ECGgen}, our model generates samples that are highly similar to the real dataset. 
According to \citet{rawshani_2019}, RBBB diagnostic characteristics include a deep and broad depression after the peak while LBBB has a wide peak and a shallow depression after the peak.
AF has a f-wave, a fibrillatory wave in the isoelectric line, and does not have a P-wave, a little uprising before the peak. 

As shown in \cref{ECGgen}, our model captures the necessary characteristics of each diagnosis, and in the figure, its significance is highlighted in red.
GRU generates similar samples regardless of the given diagnosis, whereas Transformer draws flat lines after a certain time point. 
NODE and NP tend to generate smooth ECG signals ignoring all fluctuations in ECG. 
Also, they could not synthesize necessary characteristics of the given diagnosis.  

\begin{table*}[]
\begin{center}
\caption{\label{ECGclassify}Averaged and standard deviation of classifier performance on generated ECG samples}
\resizebox{\textwidth}{!}{%
\begin{tabular}{cccccccccccccc}
\toprule
                                                                           & \multicolumn{4}{c}{Overall}             & \multicolumn{3}{c}{AF}         & \multicolumn{3}{c}{LBBB}      & \multicolumn{3}{c}{RBBB}      \\ \cline{2-14} 
                                                                           & Acc     & Recall  & Precision & F1 Score & Recall  & Precision & F1 Score & Recall  & Precision & F1 Score & Recall  & Precision & F1 Score \\ \hline
\multirow{2}{*}{GRU}                                                       & 0.448   & 0.448   & 0.419     & 0.408    & 0.143   & 0.260     & 0.184    & 0.816   & 0.475     & 0.600    & 0.383   & 0.522     & 0.441    \\
                                                                           & (0.017) & (0.017) & (0.023)   & (0.022)  & (0.035) & (0.050)   & (0.040)  & (0.015) & (0.016)   & (0.009)  & (0.056) & (0.020)   & (0.044)  \\ \hline
\multirow{2}{*}{Transformer}                                               & 0.338   & 0.338   & 0.283     & 0.239    & 0.263   & 0.481     & 0.265    & 0.002   & 0.050     & 0.004    & 0.748   & 0.317     & 0.434    \\
                                                                           & (0.022) & (0.022) & (0.028)   & (0.048)  & (0.279) & (0.102)   & (0.175)  & (0.003) & (0.103)   & (0.007)  & (0.266) & (0.016)   & (0.082)  \\ \hline
\multirow{2}{*}{NODE}                                                     & 0.488   & 0.488   & 0.496     & 0.465    & 0.240   & 0.443     & 0.309    & 0.459   & 0.587     & 0.514    & 0.765   & 0.457     & 0.572    \\
                                                                           & (0.008) & (0.008) & (0.050)   & (0.010)  & (0.034) & (0.021)   & (0.023)  & (0.038) & (0.006)   & (0.022)  & (0.024) & (0.011)   & (0.004)  \\ \hline
\multirow{2}{*}{NP}                                                        & 0.722   & 0.722   & 0.742     & 0.719    & 0.546   & \textbf{0.748}     & 0.631    & \textbf{0.757}   & 0.853     & 0.802    & \textbf{0.863}   & 0.625     & \textbf{0.725}    \\
                                                                           & (0.004) & (0.004) & (0.003)   & (0.004)  & (0.015) & (0.009)   & (0.007)  & (0.013) & (0.008)   & (0.005)  & (0.011) & (0.010)   & (0.004)  \\ \hline
\multirow{2}{*}{\begin{tabular}[c]{@{}c@{}}Fourier \\ (Ours)\end{tabular}} & \textbf{0.730}   &\textbf{0.730}   & \textbf{0.746}     & \textbf{0.734}    & \textbf{0.710}   & 0.675     & \textbf{0.691}    & 0.746   & \textbf{0.906}     & \textbf{0.818}    & 0.735   & \textbf{0.657}     & 0.693    \\
                                                                           & (0.005) & (0.005) & (0.004)   & (0.004)  & (0.056) & (0.024)   & (0.014)  & (0.019) & (0.017)   & (0.005)  & (0.040) & (0.025)   & (0.008)  \\ \hline
                                                                           \midrule
\multirow{2}{*}{Real Dataset}                                              & 0.816   & 0.796   & 0.816     & 0.805    & 0.737   & 0.756     & 0.746    & 0.777   & 0.850     & 0.812    & 0.873   & 0.841     & 0.857    \\
                                                                           & (0.005) & (0.004) & (0.005)   & (0.004)  & (0.016) & (0.014)   & (0.008)  & (0.002) & (0.007)   & (0.004)  & (0.011) & (0.009)   & (0.006) \\
                                                                           \bottomrule
\end{tabular}%
}
\end{center}

\end{table*}

We quantitatively evaluate the generated samples by using a pre-trained ECG classifier. 
The classifier is trained beforehand on the real dataset to classify the three diagnosis. 
We use total of five classifiers trained with a different parameter initialization seed.
We run the classifier on our generated samples in order to confirm whether our samples are classified to their given diagnosis. 
We report our results in \cref{ECGclassify}.

Our model outperforms all other baselines for diagnosis-averaged overall scores, showing notable performance in AF.
We speculate the reason for this improved performance is that our model is the only model that can synthesize f-wave, a main feature of AF, while other models fail to generate such fine oscillations.


\section{Conclusion}
\label{conclusion}
In this work, we introduce a new Fourier-based model architecture that can generate periodic signals.
Conventional sequential models collapse, diverge, or ignore subtle details even if they modeled the periodicity successfully. 
In contrast, our model outperforms all baseline models in all tasks across two datasets, showing stable and refined results.
We plan to extend out work to more diverse signals with various periods, since we only examined the samples with a period of 1.
We believe the periodic structure inherited by the Fourier series will enable new possibilities to model various periodic signals even beyond bio-signals in the future.

\newpage
\bibliographystyle{plainnat}
{\small
\bibliography{references}
}


\newpage
\appendix
\section{Dataset}
\label{Appen:DataPreprocess}
\subsection{Toy Dataset}
The number of samples in each class is 20,000 in train set and 5,000 for validation and test sets. 
We have total 80,000 samples in train set and 20,000 for validation and test sets.
Each sample lasts for three seconds and has total of 500 time steps. 
We added Gaussian noise sampled from $\mathcal{N}(\mathbf{0}, diag(0.3))$ in all samples.

\subsection{Physionet 2021}
We filtered out those which sampling rate is not 500Hz and those that last less than 10 seconds, or longer than 20 seconds.  
We utilized data between 5 to 10 seconds because the majority of samples include extreme noise up to 5 second and stabilize afterward. With using ECG R-peak detectors\footnote{https://github.com/berndporr/py-ecg-detectors} \citep{christov2004real}, we maintain samples that have one or two detected R peaks within one second. 
This process filters out extremely noisy samples, such as samples with all zero values or samples that do not show any dominant QRS peaks but rather a random noise. 
The total number of diagnosis label in final preprocessed dataset is shown in Table \ref{numoflabel}.

\begin{table*}[h]
\begin{center}
\caption{\label{numoflabel} Number of diagnosis labels in final dataset}
\begin{tabular}{c|c}
\toprule
     & \# of Samples \\ \hline
RBBB & 19,425          \\
LBBB & 5,270          \\
AF   & 11,415          \\ \hline
Total &   36,110            \\
\bottomrule
\end{tabular}
\end{center}
\end{table*}

\section{Experiment Details}
\label{Appen:Experiments}
\subsection{Model Architecture}

\begin{table*}[h]
\begin{center}
\caption{\label{Encoder} Encoder architecture}
\begin{tabular}{cc}
\toprule
Layer   & Layer Information                                                    \\ \hline
Layer 1 & Conv(\# of output channels=256, Kernel=3, Stride=1), MaxPool(2), SiLU \\
Layer 2 & Conv(\# of output channels=256, Kernel=3, Stride=1), MaxPool(2), SiLU \\
Layer 3 & Conv(\# of output channels=256, Kernel=3, Stride=1), MaxPool(2), SiLU \\
Layer 4 & Conv(\# of output channels=256, Kernel=3, Stride=1), MaxPool(2), SiLU \\
Layer 5 & Conv(\# of output channels=128, Kernel=3, Stride=1) \\                  
\bottomrule
\end{tabular}
\end{center}
\end{table*}
\begin{table*}[h]
\begin{center}
\caption{Decoder architecture}
\begin{tabular}{ccccc}
\toprule
               & Latent Dimension & Hidden Dimension & \# of Layers & \# of Multi-Head Attention \\ \hline
Fourier (Ours) & 128              & 256              & 6            & -                          \\
NP             & 128              & 256              & 6            & -                          \\
NODE          & 128              & 256              & 6            & -                          \\
Transformer    & 128              & 256              & 2            & 4                          \\
GRU            & 128              & 256              & 2            & -                         
\\
\bottomrule
\end{tabular}
\end{center}
\end{table*}

\subsection{Hyperparameters}
We share a fixed hyperparameter set for Toy Datset and Physionet2021 as follows.
\begin{itemize}
    \item dropout = 0.1
    \item lr = 1e-4
    \item batch size = 512
    \item $\beta = 1$ for Toy Dataset, $\beta=30$ for Physionet2021
\end{itemize}
\subsection{Experiment environments}
We train our models on NVIDIA GeForce RTX 3090. Also, CUDA version is 11.1 and torch version is 1.8.1.

\section{Additional Experiment Results}
\label{Appen:AdditionalExperiments}
\begin{figure}[ht]
\centering
  \includegraphics[width=\linewidth]{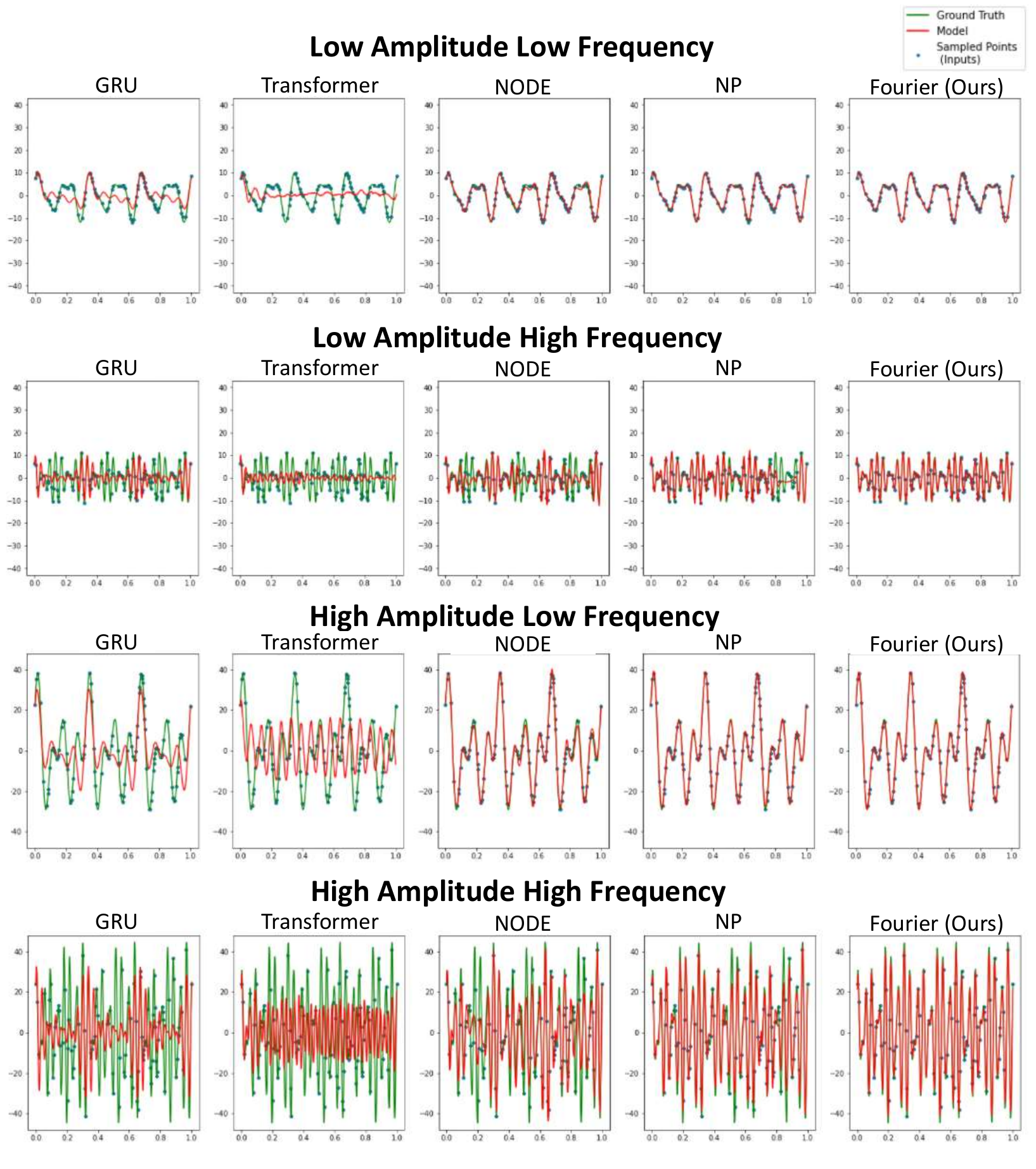}
  \caption{\label{sin_recon}Illustrated examples of reconstruction and imputation on toy dataset. Green line indicates the ground truth and blue points are sampled inputs. Red line is the model's prediction.}
\end{figure}

\begin{figure}[ht]
\centering
  \includegraphics[width=0.65\linewidth]{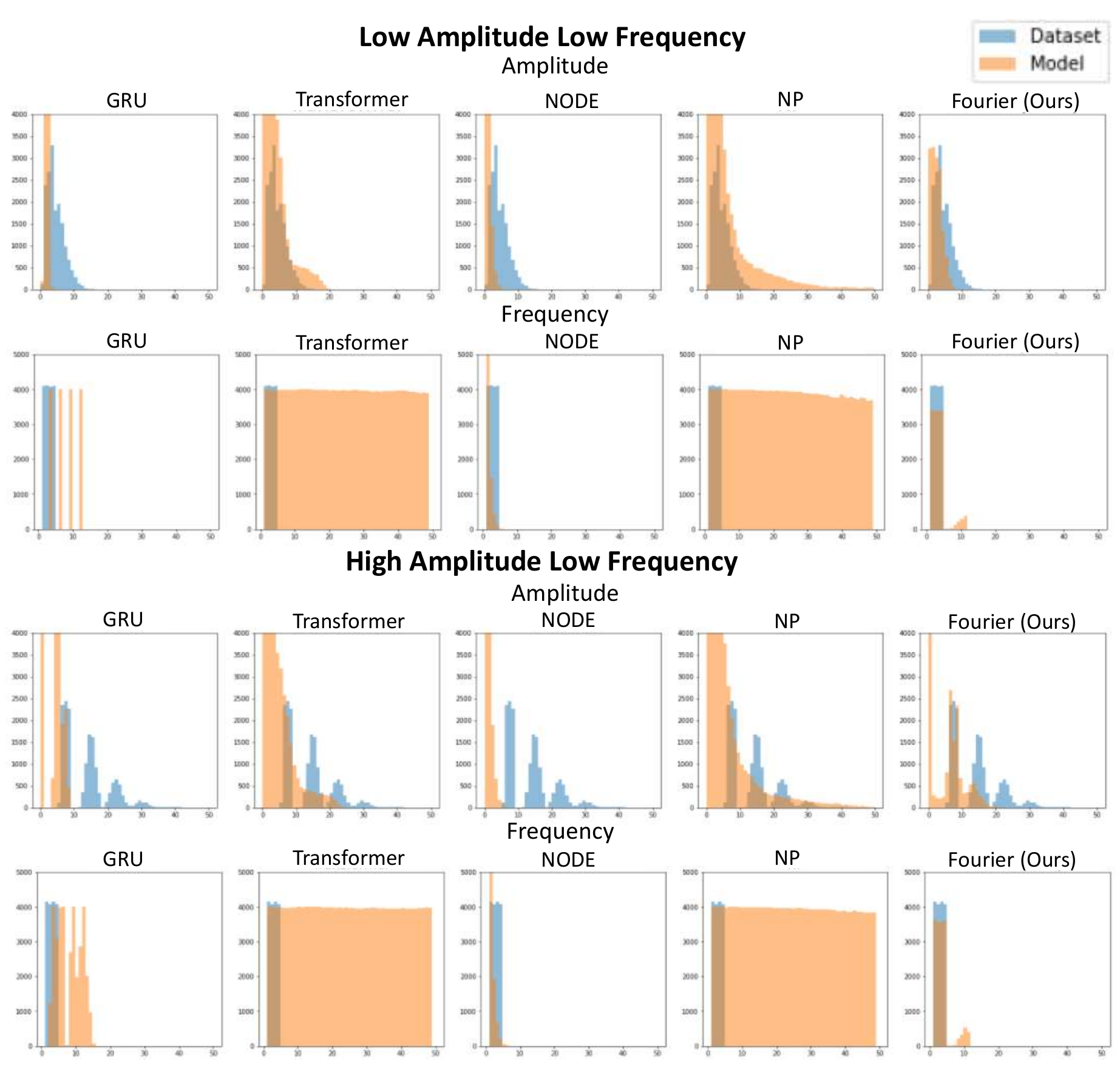}
  \caption{\label{sin_his_appen} Fourier anaylsis in `Low Amplitude \& Low Frequency' and `High Amplitude \& Low Frequency'}
\end{figure}

\begin{figure}[ht]
    \centering
    \includegraphics[width=0.65\linewidth]{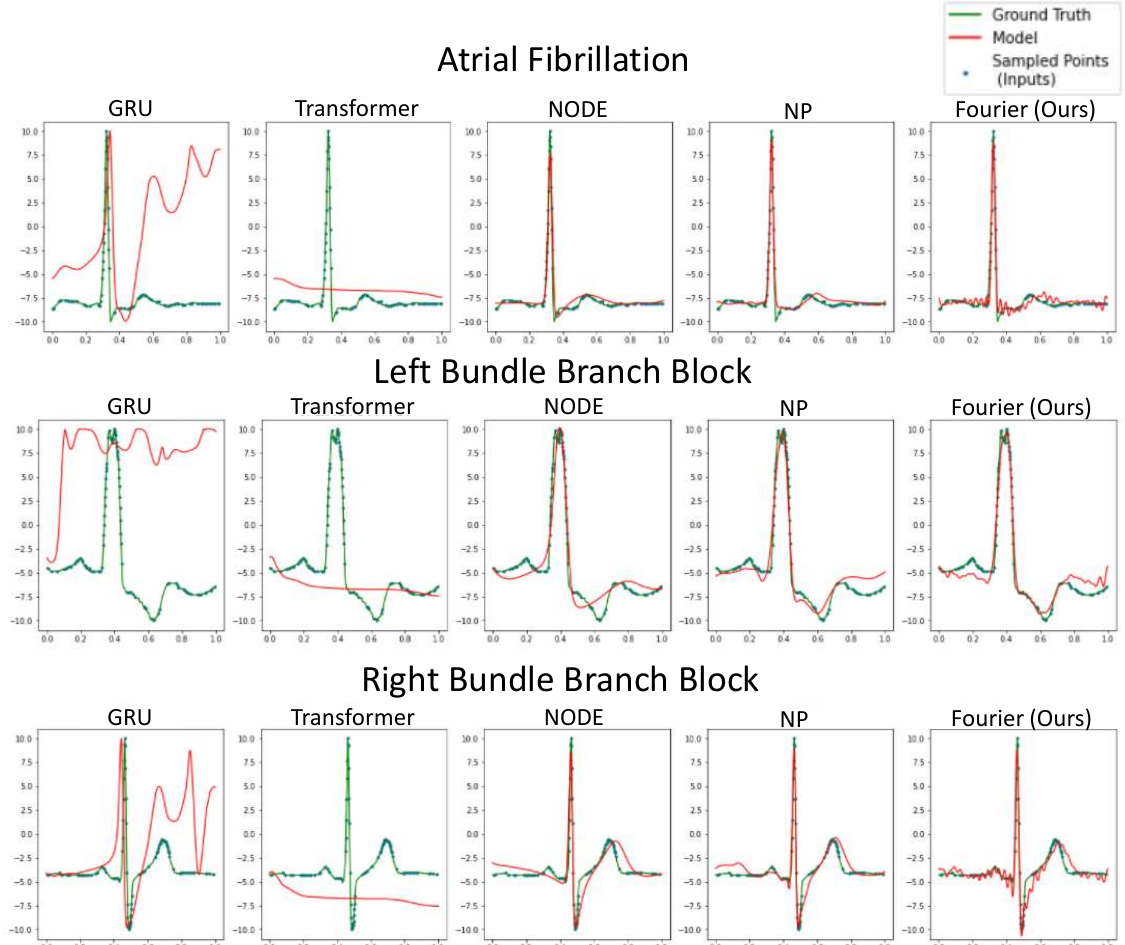}
    \caption{Illustrated examples of reconstruction and imputation on Physionet2021. Green line is the ground truth and blue points are the sampled inputs. Red line is the model's prediction.}
    \label{ECG_recon}
\end{figure}

\section{Experiment Result without Sampling}
\label{Appen:Experimentwosampling}
This section describes the results from the experiments without sampling. That is, the model receives all available input sequences.

\subsection{Experiment Results of Toy Dataset}

\begin{figure}[ht]
    \centering
    \includegraphics[width=0.8\linewidth]{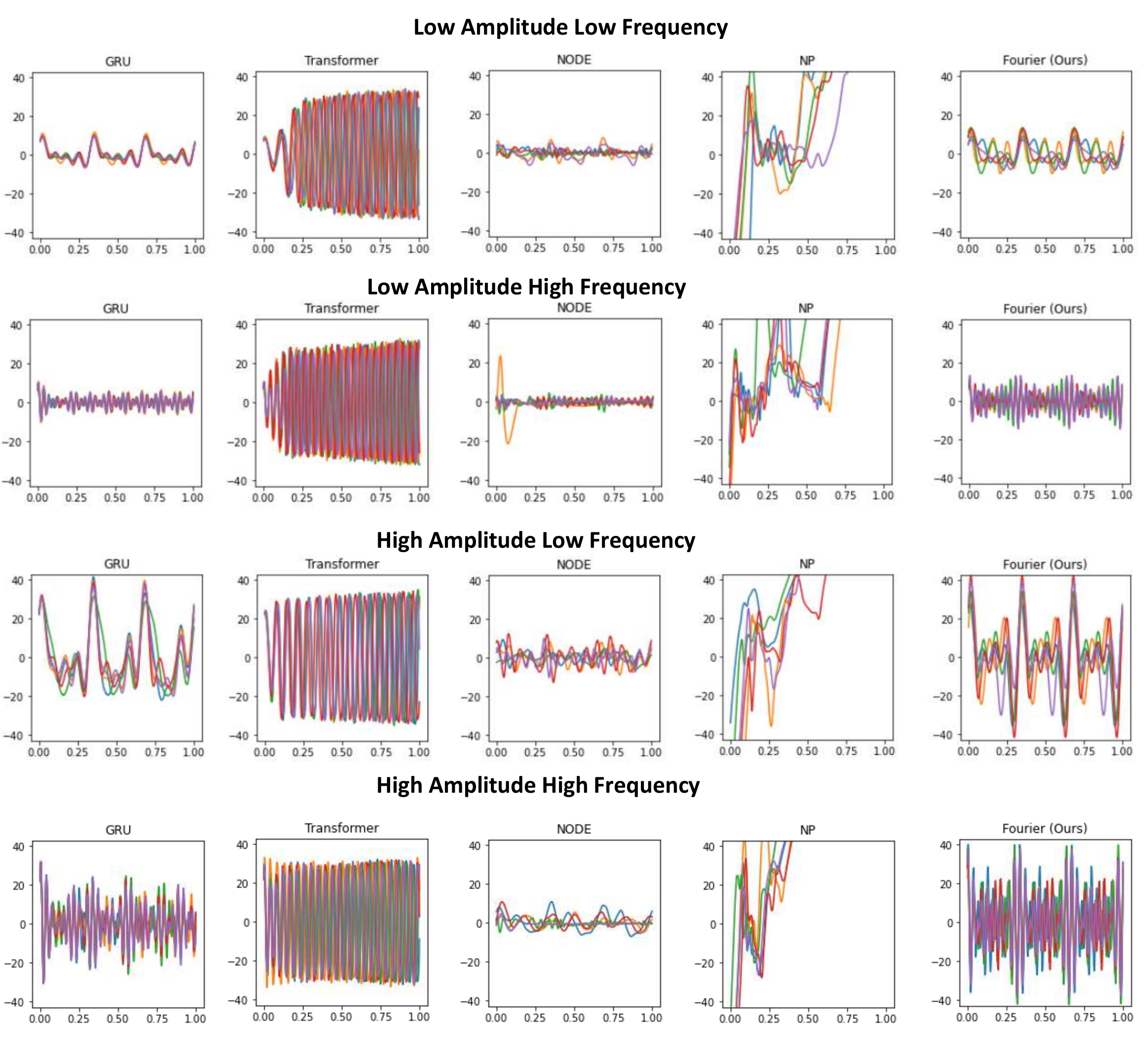}
    \caption{Conditionally generated samples in toy dataset without sampling. We draw five generated samples for each model plotted in different colors.}
    \label{fig:sin_gen_diverse_wosampling}
\end{figure}

\begin{figure}[ht]
    \centering
    \includegraphics[width=0.8\linewidth]{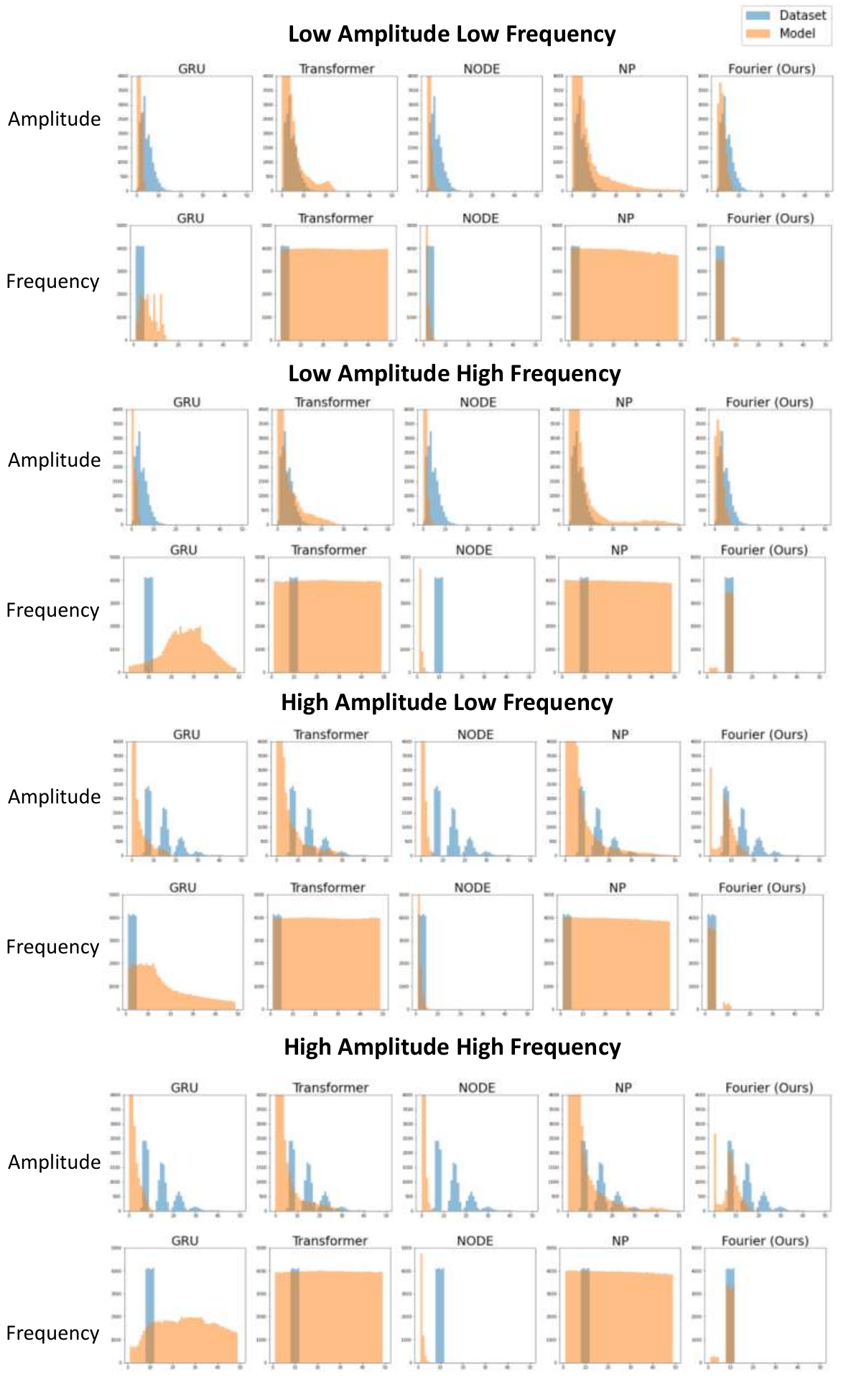}
    \caption{Fourier series analysis on conditionally generated samples without sampling. Blue color represents the original dataset whereas orange color represents each model.}
    \label{fig:sin_hist_wosampling}
\end{figure}

Generated samples on each amplitude-frequency classes are visualized in \cref{fig:sin_gen_diverse_wosampling}. 
The overall characteristics are similar to \cref{sin_gen_diverse} except that GRU can model more diverse samples. 
we report histograms on amplitude and frequency based on Fourier series analysis in \cref{fig:sin_hist_wosampling}.
The overall distributions are alike to \cref{sin_hist} but GRU covers more wide space in frequency.

\subsection{Experiment Results of Electrocardiogram}
\begin{figure}[ht]
    \centering
    \includegraphics[width=\linewidth]{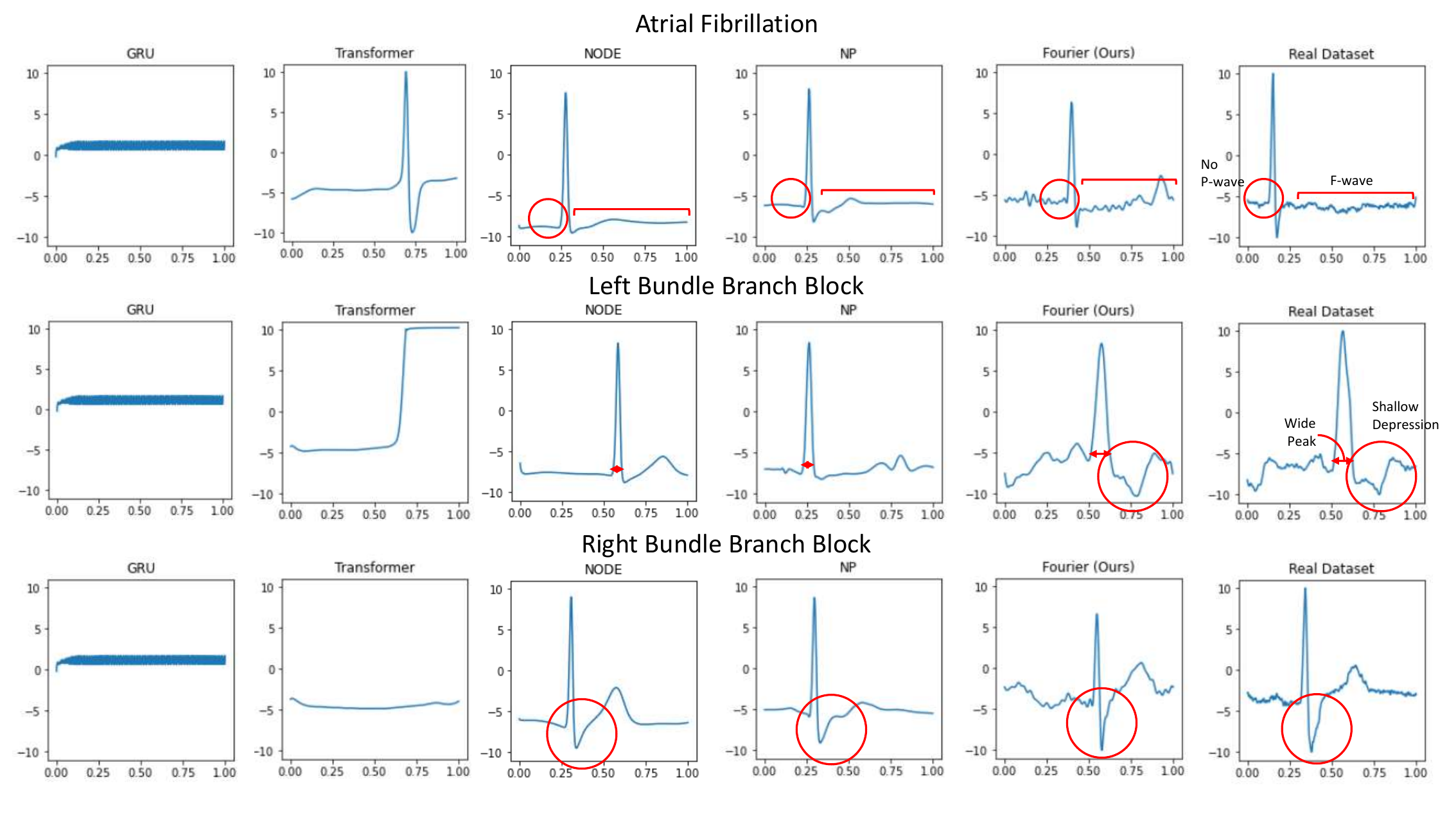}
    \caption{Conditionally generated ECG samples for each diagnosis. Main characteristics for each diagnosis and corresponding spots in generated samples are marked in red.}
    \label{fig:ecg_gen_wo_sampling}
\end{figure}
\begin{table*}[h]
\begin{center}
    
\caption{\label{ECGclassify_wo_sampling}Averaged and standard deviation of classifier performance on generated ECG samples}
\resizebox{\textwidth}{!}{%

\begin{tabular}{cccccccccccccc}
\toprule
                                & \multicolumn{4}{c}{Overall}                                       & \multicolumn{3}{c}{AF}                           & \multicolumn{3}{c}{LBBB}                         & \multicolumn{3}{c}{RBBB}                         \\ \cline{2-14} 
                                & Acc            & Recall         & Precision      & F1 Score       & Recall         & Precision      & F1 Score       & Recall         & Precision      & F1 Score       & Recall         & Precision      & F1 Score       \\ \hline
\multirow{2}{*}{GRU}            & 0.333          & 0.333          & 0.110          & 0.167          & 0.000          & 0.000          & 0.000          & 0.000          & 0.000          & 0.000          & \textbf{1.000} & 0.330          & 0.500          \\
                                & (0.000)        & (0.000)        & (0.000)        & (0.000)        & (0.000)        & (0.000)        & (0.000)        & (0.000)        & (0.000)        & (0.000)        & (0.000)        & (0.000)        & (0.000)        \\ \hline
\multirow{2}{*}{Transformer}    & 0.557          & 0.557          & 0.552          & 0.446          & 0.000          & 0.200          & 0.000          & 0.670          & \textbf{1.000} & 0.714          & \textbf{1.000} & 0.455          & 0.623          \\
                                & (0.152)        & (0.152)        & (0.128)        & (0.158)        & (0.000)        & (0.447)        & (0.001)        & (0.456)        & (0.001)        & (0.409)        & (0.000)        & (0.070)        & (0.069)        \\ \hline
\multirow{2}{*}{NODE}           & 0.531          & 0.531          & 0.574          & 0.515          & 0.306          & 0.476          & 0.371          & 0.454          & 0.773          & 0.572          & 0.832          & 0.471          & 0.601          \\
                                & (0.010)        & (0.010)        & (0.006)        & (0.013)        & (0.034)        & (0.020)        & (0.023)        & (0.021)        & (0.009)        & (0.015)        & (0.028)        & (0.011)        & (0.007)        \\ \hline
\multirow{2}{*}{NP}             & 0.709          & 0.709          & \textbf{0.751} & 0.710          & 0.548          & \textbf{0.734} & 0.626          & 0.698          & 0.932          & 0.798          & 0.883          & 0.588          & \textbf{0.734} \\
                                & (0.012)        & (0.012)        & (0.004)        & (0.013)        & (0.046)        & (0.019)        & (0.024)        & (0.013)        & (0.011)        & (0.006)        & (0.022)        & (0.022)        & (0.068)        \\ \hline
\multirow{2}{*}{Fourier (Ours)} & \textbf{0.711} & \textbf{0.711} & 0.741          & \textbf{0.716} & \textbf{0.779} & 0.625          & \textbf{0.692} & \textbf{0.714} & 0.954          & \textbf{0.816} & 0.642          & \textbf{0.645} & 0.642          \\
                                & (0.011)        & (0.011)        & (0.006)        & (0.011)        & (0.053)        & (0.032)        & (0.005)        & (0.035)        & (0.011)        & (0.020)        & (0.052)        & (0.030)        & (0.016)        \\ \hline
                                \midrule
\multirow{2}{*}{Real Dataset}   & 0.816          & 0.796          & 0.816          & 0.805          & 0.737          & 0.756          & 0.746          & 0.777          & 0.850          & 0.812          & 0.873          & 0.841          & 0.857          \\
                                & (0.005)        & (0.004)        & (0.005)        & (0.004)        & (0.016)        & (0.014)        & (0.008)        & (0.002)        & (0.007)        & (0.004)        & (0.011)        & (0.009)        & (0.006)       \\
                                \bottomrule
\end{tabular}%
}
\end{center}
\end{table*}
Generated ECG samples from each model are visualized in \cref{fig:ecg_gen_wo_sampling}.
GRU produces the same samples across all diagnoses. 
We report the mean and standard deviation of classifier performance on generated ECG samples in \cref{ECGclassify_wo_sampling}. 
Similar to \cref{ECGclassify}, our model shows decent performance in diagnosis-averaged overall score and impressive performance in AF. 
Since GRU generates the same samples regardless of diagnosis, recall in RBBB is 1 and all scores in other diagnoses scores are 0. 
Likewise, Transformer synthesize AF samples as RBBB, causing recall in RBBB 1 and scores in AF nearly 0.

\end{document}